\documentclass[10pt,twocolumn,letterpaper]{article}

\usepackage{iccv}
\usepackage{times}
\usepackage{epsfig}
\usepackage{graphicx}
\usepackage{amsmath}
\usepackage{amssymb}

\usepackage[pagebackref=true,breaklinks=true,letterpaper=true,colorlinks,bookmarks=false]{hyperref}

\iccvfinalcopy 

\def\iccvPaperID{31} 

\ificcvfinal\fi

\begin{document}

\title{360° from a Single Camera:\\A Few-Shot Approach for LiDAR Segmentation}

\author{
Laurenz Reichardt$^{1}$ \hfill
Nikolas Ebert$^{1, 2}$ \hfill
Oliver Wasenm\"uller$^{1}$ \hfill
\\
$^{1}$Mannheim University for Applied Science, Germany\\
$^{2}$RPTU Kaiserslautern-Landau, Germany\\
\tt\small 
l.reichardt@hs-mannheim.de,
n.ebert@hs-mannheim.de, 
o.wasenmueller@hs-mannheim.de
}

\maketitle
\ificcvfinal\thispagestyle{empty}\fi

\begin{abstract}
Deep learning applications on LiDAR data suffer from a strong domain gap when applied to different sensors or tasks. In order for these methods to obtain similar accuracy on different data in comparison to values reported on public benchmarks, a large scale annotated dataset is necessary. However, in practical applications labeled data is costly and time consuming to obtain. Such factors have triggered various research in label-efficient methods, but a large gap remains to their fully-supervised counterparts. Thus, we propose ImageTo360, an effective and streamlined few-shot approach to label-efficient LiDAR segmentation. Our method utilizes an image teacher network to generate semantic predictions for LiDAR data within a single camera view. The teacher is used to pretrain the LiDAR segmentation student network, prior to optional fine-tuning on 360° data. Our method is implemented in a modular manner on the point level and as such is generalizable to different architectures. We improve over the current state-of-the-art results for label-efficient methods and even surpass some traditional fully-supervised segmentation networks.
\end{abstract}

\section{Introduction}
\begin{figure}[t]
    \centering
    \includegraphics[width=\columnwidth]{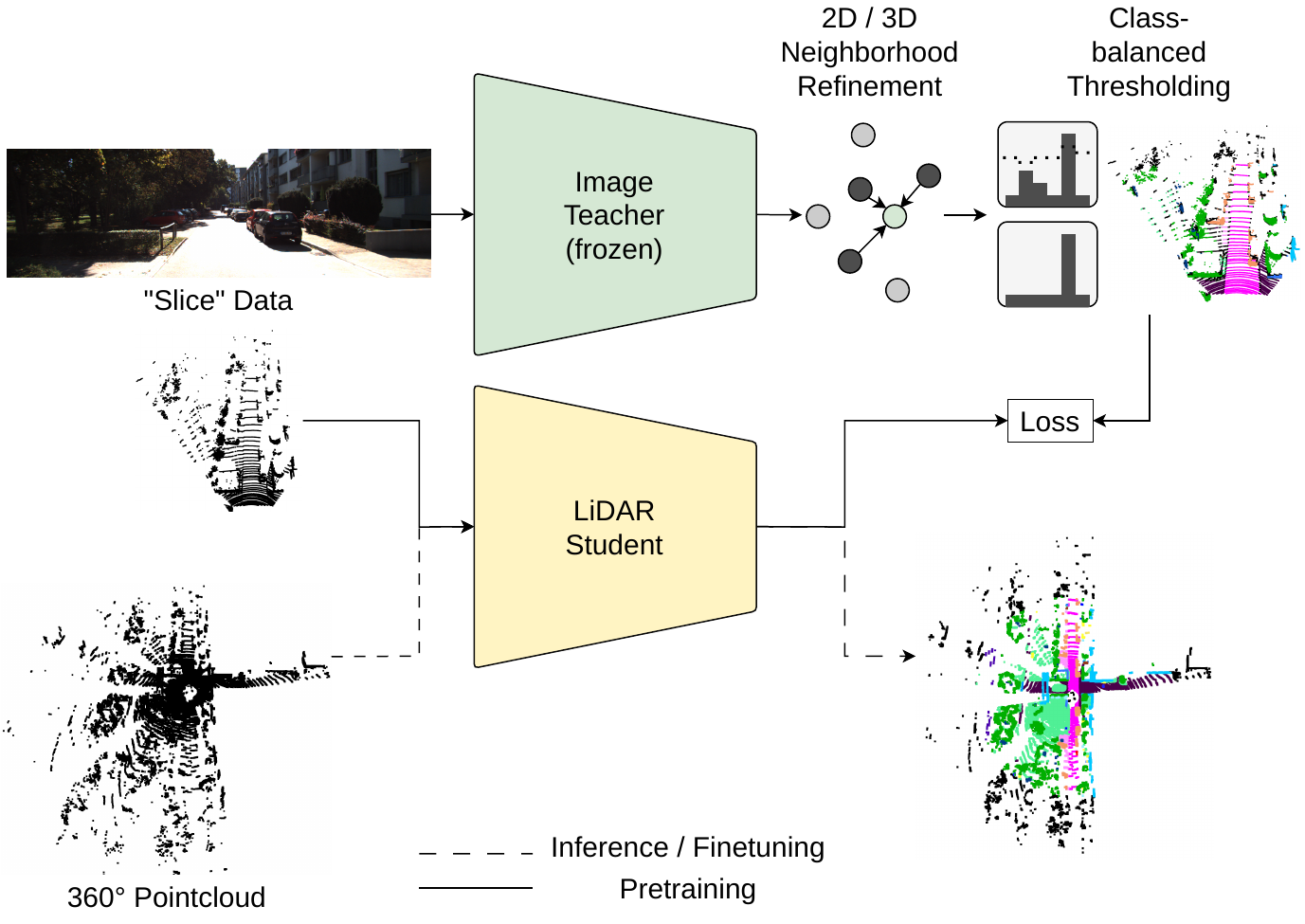}
    \caption{Our few-shot method ImageTo360. A frozen image teacher is used to pretrain the LiDAR student on single camera predictions ("Slice" Data). Later finetuning and inference is performed using the entire 360° of LiDAR data, without image data. During pretraining / finetuning the loss of the respective LiDAR student method is used, following the original implementations.}
    \label{ThisMethod}
\end{figure}
\subsection{Label Efficient LiDAR Segmentation}
Recent advancements in the application of deep learning methods for LiDAR perception have yielded impressive results on public benchmarks. It is desirable for these methods to perform consistently across different devices and specifications. However, in practice the heterogeneous characteristics of LiDAR sensors (field of view, number of beams, rotational frequency, etc.) result in substantial variations in data. This leads to a decline in performance when deep learning methods are applied to different sensors or tasks. The severity of this sensor domain problem is unique to 3D pointclouds and has prevented the widespread adoption of pretrained feature extraction backbone networks, even between different tasks of the same dataset.

As a result, the practical application of the methods require large annotated datasets in order for them to perform on par with their public benchmark counterparts \cite{DomainAdaptionSurvey}. This has led to research in efficient labeling \cite{LATTE, SALT} and Label-efficient training as natural fits to counteract the extensive efforts and costs necessary to obtain such datasets. While recent improvements in label-efficient training show promise, this field still requires more research, as most methods under-perform compared to their fully-supervised counterparts.

Building on the readily available camera images in autonomous driving and robotics we propose ImageTo360 for LiDAR semantic segmentation. In this paper we present:
\begin{itemize}
    \item A streamlined, yet effective and practically viable few-shot learning approach for 360° LiDAR semantic segmentation from a single camera view. Image data is not required at inference time
    \item The study of effective methods to improve the quality of 2D information for 3D pretraining
    \item We show that information from a limited view is sufficient to train a network for 360° data.
    \item A comprehensive analysis in the field of label-efficient LiDAR semantic segmentation, on the SemanticKITTI \cite{SemanticKITTI} benchmark, showing the effectiveness of our approach. We achieve state-of-the-art results with only 1\% data, over comparable label-efficient methods.
\end{itemize}

\section{Related Work}
The continuous and geometrically complex nature of LiDAR data makes the labeling process especially time-consuming and costly. This is exasperated by drift issues and memory requirements, limiting labeling on reconstructed scenes \cite{SemanticKITTI}.

Such factors have lead to research on training in a weakly supervised manner. A common approach is the integration of region proposals, e.g. through clustering, during the training process \cite{FusionLESS, OneClick}. ScribbleKITTI \cite{ScribbleKITTI} directly trains with weak scribble annotations, combining various methods such as self-training, multi-scale voxel class distribution information and the mean teacher framework \cite{MeanTeacher}. Besides its susceptibility to various hyperparameters, ScribbleKITTI requires a large computational budget. Furthermore, generalization to different network architectures requires individual adaptions. ScribbleKITTI scales from weakly annotated data, which requires additional human effort.

LaserMix \cite{LaserMix} combines compositional mixing of unlabeled and fully labeled data in a semi-supervised approach. However, the angular partitioning of their composition approach is tuned to the hardware of the KITTI dataset, which may impact performance with different LiDAR sensors. LiDAL \cite{LidalInterframeUncertainty} integrates selective pseudo-labeling, based on the uncertainty between multiple augmented versions of sequential LiDAR scans. Due to the use of multiple scans, their method has a high memory footprint and requires pose information. HybridCR \cite{HybridCR} combines weak supervision with consistency loss and learned augmentation.

Genova \etal~\cite{GoogleImageSeg} take an image pseudo-labeling approach at urban mapping for temporally assembled LiDAR pointclouds. Their method utilizes a series of filters which remove up to 99{\%} of points, including those of moving entities. While similar in concept, our solution fundamentally differs in that it can function on single-scan data and moving classes such as cars and pedestrians are considered. Their closed-source method was tested on a non-public dataset, making the necessary implementations for comparison infeasible.

\subsection{LiDAR Domain Adaption}
Domain adaption (DA) has been a natural candidate tackling the LiDAR cross-sensor gap. While DA is fundamentally different to label-efficient methods, the shared goal is to reduce or entirely eliminate the need for target domain data. This field is mainly split into Real\textrightarrow Real and Synthetic\textrightarrow Real domain adaption.

Some compose intermediate representations of source domain data, e.g. through voxel completion \cite{CompleteandLabel} or mesh creation \cite{MeshWorlds}, prior to sampling target domain data or using the dense data itself for training. Multi-modal methods such as xMUDA \cite{xMUDA} and ADAS \cite{ADAS} bridge LiDAR characteristics with image information. CosMix \cite{CosMix} uses a synthetic trained teacher network with compositional mixing similar to LaserMix, while SynLiDARs \cite{SynLiDAR} PCT module learns to reconstruct synthetic pointclouds in the appearance of the target domain. 

A reoccurring issue of DA methods is the evaluation on a limited number of overlapping classes between different public datasets \cite{xMUDA, DsCML, ADAS, GIPSO-ClassOverlap, CompleteandLabel, MeshWorlds, GatedUDA-ClassOverlap}. There is no consensus on mapping overlapping labels, limiting the expressiveness of results when compared to label-efficient or fully-supervised methods. Synthetic methods have the benefit that annotations can be generated to match those of the target dataset for evaluation \cite{CosMix, SynLiDAR}.

\subsection{Pointcloud Pretraining}
Advances in image self-supervised pretraining \cite{MAE, MoCoV3} have kindled research in adapting these concepts to pointclouds. Variants of masked pretraining use reconstruction \cite{PointBERT, PointMAE}, occlusion completion \cite{OcclusionCompletion}, or occupancy prediction \cite{VoxelMAE}, but are specialized towards specific network architectures or limited to data generated from 3D models. Other methods of pointcloud pretraining include contrastive methods \cite{PointContrast, SelfSupervisedContast}. However, the application of these methods has been limited to indoor pointclouds and to impact fully-supervised training, not in the context of label-efficient training.

Pointcloud pretraining also includes multi-modal strategies. Janda \etal~\cite{ContrastiveImageTraining} add image depth estimation features to their contrastive method. Other multi-modal methods such as the SLiDR series \cite{SLiDR, SLiDR2} and CLIP2Scene \cite{Clip2Scene} utilize knowledge distillation with pretrained image, image/text backbones for 3D semantic segmentation, followed by fine-tuning with a low amount of annotated data. However, the teacher backbones in these methods lack knowledge specific to the segmenting of street accuracy, and perform on-par with LiDAR only methods.

\section{Method}
We made three main observations based on the current state-of-the-art. Methods with weak annotations \cite{OneClick, ScribbleKITTI} or semi-supervised training \cite{CosMix, LidalInterframeUncertainty} intricately combine a variety of  techniques dependent on a large number of hyperparameters. Others include image information in the training process, but constrain 3D networks into learning the 2D features of discrete pixel grids \cite{ContrastiveImageTraining} instead of in continuous space, or use a general 2D backbone \cite{Clip2Scene, SLiDR, SLiDR2}. Thirdly, most pretraining methods are specific to certain architectures or representations (voxels, range-projections, etc.)

These observations inspire us to follow a "back to the basics" approach for ImageTo360. We assume that camera data together with registered LiDAR data is readily available in autonomous driving and robotics. Our method leverages this image information in a 2D supervised teacher network pretraining of a 3D student network. Since the teacher predictions are only available within the camera field of view, we follow up by optional fine-tuning on 360° data. Our proposed few-shot method is visualized in Fig. \ref{ThisMethod}.

Pretrained image segmentation networks are readily available \cite{MMSegmentation} and can be used as "off-the-shelf" teacher networks. From a practical perspective image segmentation is a logical choice to introduce specific knowledge into label efficient LiDAR segmentation. In order to not bind the 3D student network to 2D features, our method uses high quality pseudo-labels as a general representation of segmentation knowledge.

We implement our method on the point level, as LiDAR data is first represented as pointclouds prior to transformation into other representations. As a result our method generalizes across sensors and architectures.

\subsection{2D Supervision}
We leverage the Cityscapes \cite{Cityscapes} dataset due to street scenes similar to those of SemanticKITTI \cite{SemanticKITTI}. In order to fully evaluate our method against others, it is necessary to include all SemanticKITTI classes in our analysis, rather than just reporting class-overlap scores. For this reason we fine-tune the network on a subset of the KITTI dataset, however this step is not necessary for the actual implementation of our method.

\begin{figure}[t!]
    \centering
    \includegraphics[width=0.75\columnwidth]{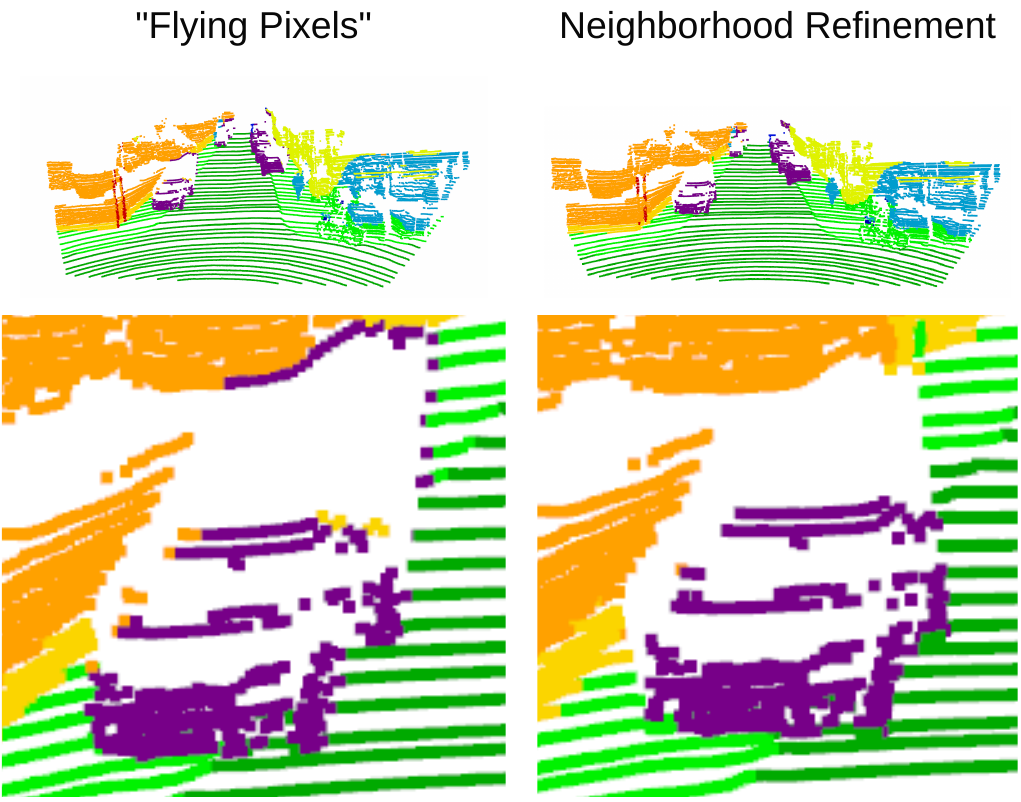}
    \caption{Image segmentation predictions projected into 3D space. Uncertainty at object edges cause "Flying Pixels" \cite{TeutscherPCD}; in this example (left image) casting car labels unto the building and vice-versa. Neighborhood refinement counteracts this issue.}
    \label{FlyingPixels}
\end{figure}

The 2D predictions are limited to the single camera field of view. When projected into 3D space, using the calibration parameters between camera and LiDAR, these predictions cover a "slice" of the 360° pointcloud (see Fig. \ref{ThisMethod}). This projection inherently has errors, caused by discrepancies in the calibration and synchronization between camera and LiDAR sensors. On top of that, neural network predictions tend to be unfocused at object borders. These discontinuous predictions cause the "Flying Pixels" bleeding effect \cite{TeutscherPCD, DVMN}, exemplified in Fig. \ref{FlyingPixels}.

While the corresponding pixels of these prediction errors are direct neighbors in 2D grids, this does not necessarily hold true when projected into 3D space. We take inspiration from post-processing methods in range projected LiDAR segmentation \cite{SalsaNext}\cite{Rangenet++} and propose 3D neighborhood refinement as a method to reintroduce continuous space information into 2D predictions. For our method, we use K-Nearest-Neighbors (KNN), but other neighborhood algorithms are also suitable.

First we consider neighborhood majority voting. However KN-Neighborhoods have varying distances and volumes. Simple majority voting would equally weight points regardless of distance. For this reason, we also consider distance-weighted voting ($1-\text{softmax}(\vec v_{d})$ with $\vec v_{d}$ as the distance vector). Priority is put on labels closer to the query point. Thirdly, we compare both methods to neighborhood confidence averaging.

\subsection{Pretraining from 2D predictions}
Pseudo-label confidence thresholding \cite{PseudoLabelThreshold} has been an established method in order to remove noisy labels due to incorrect predictions. In a multiclass context, networks tend to bias towards majority classes \cite{DebiasedHead}\cite{ClassBalancedTraining}, and simple static thresholding can remove minority classes. Class-imbalance inherently affects LiDAR street scenario data (e.g. buildings have more points than a person). We follow established methods \cite{ScribbleKITTI}\cite{GIPSO-ClassOverlap} and integrate class-balanced adaptive thresholding with the goal of further improving pseudo-label quality.

Formally, the class balanced thresholds $\vec\tau_{c}(i)$ of our method can be expressed as
\begin{equation}
\begin{split}
    \vec\tau_{c}(i) = &\frac{\sum[labels=i]}{max\left \{ \forall_{i\in c}\sum[labels=i]  \right \}} \times\\
    &(\tau_{max}-\tau_{min}) + \tau_{min}
\end{split}
\label{eq:cbt}
\end{equation}
with all dataset classes $c$, single class $i$, $\tau_{min}$ as the minimum threshold, and $\tau_{max}$ as the maximum threshold values. In short, threshold values are scaled within a confidence range (set through $\tau_{min}$ and $\tau_{max}$) for each class, depending on how many times this class occurs in comparison to the majority class. Predictions below the adaptive threshold are not used as pseudo-labels.

For pretraining, the pointclouds are cut into the fields of view of the camera image, significantly reducing training time and memory requirements. Our method can be scaled to include multiple cameras.
Optionally, fine-tuning is performed using the annotated 360° LiDAR data, without image data.
%
\section{Evaluation}
We evaluate our method ImageTo360 against related works for the spaces of fully-supervised, label-efficient and pretraining methods, as well as against domain adaption. The intention is to show the differences in performance when considering practical application and labeling efforts.

Our ablation studies and evaluation are performed using the SemanticKITTI \cite{SemanticKITTI} dataset in a single-scan context. Results are reported on evaluation sequences 08, either on the full 360° or the "slice" of data within the left camera field of view (covering ~90°). For image teacher networks we use the OpenMMLab \cite{MMSegmentation} implementation of Segformer \cite{Segformer} and DeepLabV3+ \cite{DeeplabV3+} in combination with the left image data of the dataset. For our 3D student networks we use SPVCNN \cite{SPVNas} and Cylinder3D \cite{Cylinder3D}.

For 3D student fine-tuning, we randomly sample a small percentage of the training split and use the entire 360° pointclouds and corresponding ground-truth annotations. Fine-tuning is performed according to the original methods hyperparameters, except with $0.2 \times learning\;rate,\;0.5 \times epochs$. Additionally we use translation, pointcloud mixing (similar to CutMix \cite{CutMix} and LaserMix \cite{LaserMix}), squeezing, and flipping as augmentation for both networks. Using this augmentation schedule, we reproduced the paper results of the original authors.

We do not include test data or validation data in any training steps of our method. Unless otherwise stated we do not use Test Time Augmentation (TTA) or ensembling, in order to provide a fair comparison with research which forgoes such methods. In the case where we use these methods, we utilize Greedy Soup \cite{GreedySoup} ensembling, and TTA over 12 variants of data (original, three flips, eight rotations).

\subsection{Ablation Study}
\begin{table*}[t]
\centering
\caption{\label{Ablation-KNN-Refinement} Evaluation of 3D neighborhood refinement of 2D predictions, using the KITTI validation mIOU in the left camera field of view.}
\resizebox{\textwidth}{!}{\begin{tabular}{|lccccccccccc|}
\hline
\multicolumn{1}{|c}{\textbf{Method}} & \multicolumn{11}{c|}{\textbf{Neighbors}}                                                                                                               \\
                                         & \textit{3} & \textit{5} & \textit{7} & \textit{9} & \textit{11} & \textit{13} & \textit{15} & \textit{17} & \textit{19}    & \textit{21} & \textit{23} \\ \hline
None (Segformer-baseline \cite{Segformer})                          & \multicolumn{11}{c|}{58.2}                                                                                                                             \\
KNN Majority Voting                          & 59.21      & 59.79      & 60.28      & 60.63      & 60.89       & 61.08       & 61.18       & 61.23       & 61.24          & 61.24       & 61.21       \\
KNN Distance Normalized Voting               & 59.21      & 59.79      & 60.28      & 60.64      & 60.90       & 61.09       & 61.21       & 61.27       & \textbf{61.30} & 61.29       & 61.26       \\
KNN Confidence Averaging                 & 59.20      & 59.78      & 60.23      & 60.55      & 60.78       & 60.94       & 61.02       & 61.07       & 61.07          & 61.06       & 61.02       \\ \hline
\end{tabular}}
\label{KNN}
\end{table*}
\subsubsection{Impact of Image Segmentation Network}
Firstly, we study the impact of an image teacher by pretraining the 3D student network on the predictions within the left camera field of view (\(\sim \)90°). We follow by fine-tuning on a low-label budget using 360° annotated pointclouds. The baseline comparisons train on the same finetuning data, without pretraining.

Improvements in image segmentation theoretically help in pretraining the 3D student. To test this assumption we compare Segformer as a modern transformer architecture with the traditional convolutional DeepLabV3+, as teacher networks. Segformer achieves 58.24 mIOU\% compared to the 50.88 mIOU\% of DeepLab, an improvement of +14.47\% on the left camera field of view predictions.

Results are depicted in Table \ref{ImageComparison}. Our pretraining method significantly improves over the baseline for nearly all reduced label scenarios, however the magnitude of improvement reduces with an increasing ratio of labels. The improvements of Segformer over DeepLabV3+ translate to the low label amount of the SPVCNN architecture. With Cylinder3D as the student, results are comparable across both teacher networks.
These results reveal noteworthy observations. Without fine-tuning, networks that have only ever seen slices of LiDAR data produce considerable results when predicting on 360° data. While developments in image segmentation can have an impact, results are heavily influenced by the choice of 3D student architecture. Finally, image pretraining improved fully-supervised training with both image teachers for SPVCNN, showing potential for future research.

We continue with Segformer as our teacher network for further ablation and evaluation.
\begin{table}[t!]
\centering
\caption{Comparison of Segformer \cite{Segformer} and DeepLabV3+ \cite{DeeplabV3+} mIOU as teacher networks. Evaluation is performed on the entire 360° annotated pointcloud of the KITTI validation split.}
\resizebox{\columnwidth}{!}{
\begin{tabular}{|lcccccccc|}
\hline
\multicolumn{1}{|c}{\textbf{Student / Teacher Networks}} & \multicolumn{8}{c|}{\textbf{SemanticKITTI Labels}}                                                                           \\
                                     & 0 \%  & \textit{0.3\%} & \textit{1\%}   & \textit{5\%}   & \textit{10\%}  & \textit{20\%}  & \textit{50\%}  & \textit{100\%} \\ \hline
Cylinder3D \cite{Cylinder3D} (baseline)                & -     & 27.24          & 34.38          & 48.90          & 54.91          & 59.57          & 65.44          & 66.16          \\
Cylinder3D / DeepLabV3+ \cite{DeeplabV3+}              & 46.62 & 49.76          & 54.50          & 58.29          & 58.62          & 61.44          & 64.22          & 64.88          \\
Cylinder3D / Segformer \cite{Segformer}               & 50.80 & 50.74          & 54.10          & 59.26          & 60.00          & 62.27          & 64.99          & 65.82          \\ \hline
SPVCNN \cite{SPVNas} (baseline)                    & -     & 27.25          & 32.42          & 49.18          & 59.10          & 63.38          & 65.11          & 65.81          \\
SPVCNN / DeepLabV3+                  & 45.87 & 48.53          & 56.08          & \textbf{61.96} & \textbf{63.07} & \textbf{64.38} & 65.83          & 65.94          \\
SPVCNN / Segformer                   & 52.15 & \textbf{57.30} & \textbf{59.53} & 61.71          & 62.41          & 64.19          & \textbf{66.10} & \textbf{66.44} \\ \hline
\end{tabular}
}
\label{ImageComparison}
\end{table}

\subsubsection{3D Neighborhood Refinement of 2D Predictions}
We study the effects of utilizing neighborhood refinement as a technique for reintroducing 3D information into 2D predictions of Segformer. We examine neighborhood majority label voting, distance weighted label voting, and neighborhood confidence averaging in Table \ref{KNN}. All three approaches demonstrate comparable improvements over the baseline. The use of confidence averaging offers the added benefit of avoiding one-hot labels and as such being compatible with pseudo-label thresholding. For this reason we continue with confidence averaging ($K=19$) for further ablation and evaluation, improving pseudo-label quality by +4.86\%.

\subsubsection{Class Balanced Pseudo-Label Thresholds}
We compare static and class-balanced adaptive thresholding (see Equation \ref{eq:cbt}) after neighborhood confidence averaging. The results are depicted in Table \ref{Thresholding}. We use $\tau_{max}=0.95$ as an upper limit to all adaptive threshold ranges. Threshold parameters are chosen to have a comparable percentage of points removed. Class-balanced adaptive thresholding shows a significant improvement over static thresholds, in total improving pseudo-label quality by +20.15\% over the neighborhood refined baseline (61.07 mIoU\%), while setting ~23\% of unreliable labels as unlabeled. More information is kept compared to the best static threshold and as a result we continue with adaptive thresholding ($\tau_{min}=0.8$) for further evaluation.
\begin{table}[t!]
\centering
\caption{Ablation of pseudo-label thresholding, applied after neighborhood confidence averaging, using the KITTI validation mIOU in the left camera field of view.}
\resizebox{0.9\columnwidth}{!}{
\begin{tabular}{|lcccc|}
\hline
\multicolumn{5}{|c|}{\textit{Static threshold}}                                   \\ \hline
\multicolumn{1}{|l|}{\textbf{Threshold} {$\tau$}}                & 0.80  & 0.85  & 0.90  & 0.95  \\
\multicolumn{1}{|l|}{\textbf{Point reduction \%}} & 16.83 & 20.04 & 23.63 & 28.58 \\
\multicolumn{1}{|l|}{\textbf{mIOU \%}}            & 70.95 & 71.69 & 72.19 & 72.82 \\ \hline
\multicolumn{5}{|c|}{\textit{Class-balanced threshold ($\tau_{\text{max}}$ = 0.95)}}           \\ \hline
\multicolumn{1}{|l|}{\textbf{\textbf{Min. threshold} $\tau_{min}$}}             & 0.5   & 0.6   & 0.7   & 0.8   \\
\multicolumn{1}{|l|}{\textbf{Point reduction \%}} & 15.83 & 18.31 & 20.78 & 23.41 \\
\multicolumn{1}{|l|}{\textbf{mIOU \%}}            & 68.75 & 70.95 & 72.45 & \textbf{73.38} \\ \hline
\end{tabular}
}
\label{Thresholding}
\end{table}

\subsection{Comparison with Label Efficient Methods}
\begin{table*}[t!]
\centering
\caption{Evaluation mIOU on the 360° data of the KITTI validation split. Percentages of labels signify the amount of training split data that was used during fine-tuning. The training split data within the camera field of view was used for image teacher pretraining.  We compare to the values reported by the authors. Empty "-" values signify that the original papers did not evaluate on that percentage of annotations.}
\resizebox{\textwidth}{!}{\begin{tabular}{|ccccccccccc|}
\hline
\textbf{Method}       & \textbf{Backbone Networks}                 & \textbf{Use of images} & \multicolumn{8}{c|}{\textbf{SemanticKITTI Labels}}                                                                            \\
\multicolumn{1}{|l}{} & \multicolumn{1}{l}{}                       & \multicolumn{1}{l}{}   & \textit{0\%} & \textit{0.3\%} & \textit{1\%}  & \textit{5\%}  & \textit{8\%}  & \textit{10\%} & \textit{20\%} & \textit{50\%} \\ \hline
\multicolumn{11}{|c|}{\textit{Label Efficient}}                                                                                                                                                                             \\ \hline
LiDAL \cite{LidalInterframeUncertainty}                 & SPVCNN \cite{SPVNas}                                     & no                     & -            & -              & 48.8          & 59.5          & -             & -             & -             & -             \\
ReDAL \cite{ReDAL}                 & SPVCNN                                    & no                     & -            & -              & 41.8          & 59.8          & -             & -             & -             & -             \\
HybridCR \cite{HybridCR}             & PSD \cite{PSD}                                        & no                     & -            & -              & 52.3          & -             & -             & -             & -             & -             \\
LaserMix \cite{LaserMix}              & Cylinder3D \cite{Cylinder3D}                                 & no                     & -            & -              & 50.6          & -             & -             & 60.0          & 61.9          & 62.3          \\
\textbf{ImageTo360 (ours)}         & \multicolumn{1}{r}{Segformer \cite{Segformer} / SPVCNN}     & yes                    & 52.1         & 57.3           & 59.5          & \textbf{61.7} & \textbf{62.1} & \textbf{62.4} & \textbf{64.2} & \textbf{66.1} \\
\multicolumn{1}{|l}{} & \multicolumn{1}{l}{+ TTA, Greedy Soup}     & yes                    & -            & -              & \textbf{62.9} & -             & -             & -             & -             & -             \\
\textbf{ImageTo360 (ours)}         & \multicolumn{1}{r}{Segformer / Cylinder3D} & yes                    & 50.8         & 50.7           & 54.1          & 59.2          & 59.6          & 60.0          & 62.2          & 65.0          \\ \hline
\multicolumn{11}{|c|}{\textit{Weakly Annotated}}                                                                                                                                                                            \\ \hline
ScribbleKITTI \cite{ScribbleKITTI}         & SPVCNN                                     & no                     & -            & -              & -             & -             & 61.3          & -             & -             & -             \\
ScribbleKITTI         & Cylinder3D                                 & no                     & -            & -              & -             & -             & 60.8          & -             & -             & -             \\ \hline
\multicolumn{11}{|c|}{\textit{Self-supervised Pretraining}}                                                                                                                                                                 \\ \hline
CLIP2Scene \cite{Clip2Scene}            & CLIP \cite{CLIP} / SPVCNN                              & yes                    & -            & -              & 42.6          & -             & -             & -             & -             & -             \\
SLidR \cite{SLiDR}                 & Minkowski \cite{Minkoswki}                                  & yes                    & -            & -              & 44.6          & -             & -             & -             & -             & -             \\
ST-SLidR \cite{SLiDR2}              & SwAV \cite{SwAV} / Minkowski                           & yes                    & -            & -              & 44.9          & -             & -             & -             & -             & -             \\
Contrastive Image \cite{ContrastiveImageTraining}     & ResNet18 \cite{ResNet} / Minkoswki                       & yes                    & \multicolumn{8}{c|}{42.0 (Percentage not stated)}                                                                             \\ \hline
\end{tabular}}
\label{EvalTable}
\end{table*}
We compare our method ImageTo360 with research from the spaces of weakly-supervised, few-shot, and semi-supervised training, as well as self-supervised pretraining. All compared methods in this evaluation rubric share the goal of label-efficient training, fine-tuning on a subset of annotated KITTI data. This rubric excludes efficient annotation methods, which focus on reducing the labeling effort, instead of the data needed. 
It is common in this field of research to evaluate on the validation set, without ranking on the test benchmark. Also, there is no agreed-upon percentage of labels for evaluation, so we've reported our results (see Table \ref{EvalTable})  using a variety of percentages.

As the most common reported metric, 1\% labels can be considered the most indicative of performance. Comparing amongst the same 3D backbone, ImageTo360 improves +17.58\% over LaserMix.  We apply TTA and ensembling to our best 1\% network to show the full potential of our method, improving over next-best HybridCR by +20.26\%. We even surpass ScribbleKITTI with its 8\% labeled points. Taken to the extremes by fine-tuning on just 57 annotated samples (0.3\%), we outperform all comparable methods using 1\% of data, and achieve similar results to those using 5\% of data. 

Current image backbone pretraining methods usually transfer 2D features to 3D networks. Considering the gap in performance to label-efficient LiDAR-only methods, this multi-modal knowledge is not necessarily beneficial. Our method differs by introducing the specific knowledge of an off-the-shelf segmentation backbone, outperforming all other multi-modal pretraining methods, even without any fine-tuning. From a application standpoint, segmentation backbones are readily available and lead to a considerable performance increase.

\subsection{Comparison with domain adaption methods}
From a data-perspective, evaluating our ImageTo360 to LiDAR domain adaption (DA) methods is biased. Even with teacher networks trained on a source domain, cross-sensor DA without target domain annotations remains a challenge. Regardless of our method using a modest amount of data, we believe that this comparison is worthwhile when considering practical applications, even more so for safety critical applications where accuracy is of great importance.

Most DA methods evaluate on the class-overlap between public benchmarks. Currently, there is no consensus amongst authors on the class-mapping. This makes the comparison of DA methods to each other challenging. However, the performance margins of each DA method to our method as a reference can give indication on what strategy performs best. We reevaluate our best 1\% method, according to the class-mapping of each compared method. The results are depicted in Table \ref{UDA}.
The margins of improvement point to the benefits of image information in DA. On AudiA2D2 \cite{AudiA2D2} \textrightarrow KITTI our method performs +41.10\% better. The margin of improvement is much greater when considering the best method without image data (SynLiDAR \textrightarrow KITTI with +95.34\%). Both results put into view the trade-off between the efforts of annotating a limited amount of data or choosing domain-adaption approaches.
\begin{table}[t!]
\centering
\caption{Comparison of our best 1\% method to domain adaption. We reevaluate using the label mappings provided by the authors. Evaluation on the KITTI validation sequence.}
\resizebox{\columnwidth}{!}{
\begin{tabular}{|llcc|}
\hline
\multicolumn{1}{|c}{\textbf{Method}} & \multicolumn{1}{c}{\textbf{Backbones}} & \textbf{Use of images} & \textbf{mIOU \%} \\ \hline
\multicolumn{4}{|c|}{\textit{A2D2 \cite{AudiA2D2} to KITTI \cite{SemanticKITTI} (10 classes)}}                                                                 \\ \hline
xMUDA \cite{xMUDA}                                & SparseVoxel \cite{SparseVoxel} + ResNet34 \cite{ResNet}                 & yes                    & 49.1             \\
DsCML \cite{DsCML}                               & SparseVoxel + ResNet34                 & yes                    & 52.4             \\
ADAS \cite{ADAS} + DsCML                         & SparseVoxel + ResNet34                 & yes                    & 54.3             \\
\textbf{ImageTo360 1\% (ours)}                        & SPVCNN \cite{SPVNas} / Segformer \cite{Segformer}                      & yes                    & \textbf{76.62}   \\ \hline
\multicolumn{4}{|c|}{\textit{Synth4D \cite{GIPSO-ClassOverlap} (synthetic) to KITTI (7 classes)}}                                                   \\ \hline
GIPSO \cite{GIPSO-ClassOverlap}                               & MinkUNet \cite{Minkoswki}                               & no                     & 40.24            \\
\textbf{ImageTo360 1\% (ours)}                        & SPVCNN + Segformer                      & yes                    & \textbf{83.25}   \\ \hline
\multicolumn{4}{|c|}{\textit{NuScenes \cite{NuScenes} to KITTI (10 classes)}}                                                             \\ \hline
Complete \& Label \cite{CompleteandLabel}                    & SparseVoxel                            & no                     & 33.7             \\
\textbf{ImageTo360 1\% (ours)}                        & SPVCNN + Segformer                      &                        & \textbf{79.03}   \\ \hline
\multicolumn{4}{|c|}{\textit{NuScenes to KITTI (11 classes)}}                                                             \\ \hline
Fake it, Mix it \cite{MeshWorlds}                      & Cylinder3D \cite{Cylinder3D}                             & no                     & 34.3             \\
\textbf{ImageTo360 1\% (ours)}                        & SPVCNN + Segformer                      & yes                    & \textbf{77.62}   \\ \hline
\multicolumn{4}{|c|}{\textit{NuScenes to KITTI (11 classes, different mapping)}}                                          \\ \hline
GatedAdapters \cite{GatedUDA-ClassOverlap}                        & SalsaNext \cite{SalsaNext}                              & no                     & 23.5             \\
\textbf{ImageTo360 1\% (ours)}                        & SPVCNN + Segformer                      & yes                    & \textbf{72.23}   \\ \hline
\multicolumn{4}{|c|}{\textit{SynLiDAR \cite{SynLiDAR} (synthetic) - KITTI (all classes)}}                                                 \\ \hline
SynLiDAR                             & PCT \cite{SynLiDAR} / Minkowski                        & no                     & 27.0             \\
CoSMix \cite{CosMix}                               & Minkowski                              & no                     & 32.2             \\
\textbf{ImageTo360 1\% (ours)}                    & SPVCNN + Segformer                          & yes                    & \textbf{62.9}    \\ \hline
\end{tabular}
}
\label{UDA}
\end{table}

\subsection{Comparison with fully-supervised methods}
For our comparison with fully-supervised semantic segmentation methods we use our best 1\% network (Segformer / SPVCNN + TTA, ensembling). We follow standard practice and compare to methods using 100\% human annotated data on the KITTI test benchmark. Our method is at a natural disadvantage, even more so considering we do not fine-tune with any validation data or use semi-supervised training on the test split.

The results on the KITTI benchmark are depicted in Table \ref{FullySupervised} and can be interpreted two-fold. While ImageTo360 outperforms comparable methods, few-shot training is still an open field of research. A gap remains to current state-of-the-art fully supervised networks, with 2DPASS \cite{2DPass} performing +26.34\% better than our method. But with considerable less human effort and annotation costs, our few-shot method outperforms traditional fully-supervised methods such as SqueezeSegV3 \cite{SqueezeSeq}. We encourage future label-efficient research to also upload their results on the public benchmark.
\begin{table}[t!]
\centering
\caption{Evaluation of our method compared to fully-supervised methods on the KITTI single-scan semantic segmentation benchmark.}
\begin{tabular}{|llc|}
\hline
\multicolumn{1}{|c}{\textbf{Method}} & \multicolumn{1}{c}{\textbf{Image training}} & \textbf{mIOU \%} \\ \hline
2DPASS \cite{2DPass}                               & \multicolumn{1}{c}{yes}                     & 72.9             \\
Point-Voxel KD \cite{PointVoxelKD}                       &                                             & 71.2             \\
Cylinder3D \cite{Cylinder3D}                           &                                             & 67.8             \\
SPVNAS \cite{SPVNas}                               &                                             & 66.4             \\
JS3C-Net \cite{JS3C}                             &                                             & 66.0             \\
SalsaNext \cite{SalsaNext}                           &                                             & 59.5             \\
KPConv \cite{KPConv}                               &                                             & 58.8             \\
\textbf{ImageTo360 1\% (ours)}                   & \multicolumn{1}{c}{yes}                     & 57.7             \\
SCSSnet \cite{SCSS}                              &                                             & 57.6             \\
SqueezeSegV3 \cite{SqueezeSeq}                        &                                             & 55.9             \\
3D-MiniNet \cite{3DMiniNet}                           &                                             & 55.8             \\
RangeNet53++ \cite{Rangenet++}                         &                                             & 52.2             \\ \hline
\end{tabular}
\label{FullySupervised}
\end{table}
\vspace{-4mm}

\section{Conclusion}
In this paper, we present ImageTo360, a streamlined approach to few-shot LiDAR semantic segmentation using an image teacher network for pretraining. Our method is designed in a modular manner and at the point level, meaning it can be applied across different 3D network architectures and that components can be exchanged with further developments in deep learning.
We evaluated against other label-efficient methods, producing state-of-the-art results. With practical applications in mind, we expanded our evaluation against domain adaption and fully-supervised methods, showing the large performance gaps between the different fields. The results show that image data of a comparatively low cost camera is sufficient for 3D networks to generate remarkable results on 360° LiDAR data, even outperforming traditional fully-supervised methods.
\section*{Acknowledgement}
This work was funded by the German Federal Ministry for Economic Affairs and Climate Action (BMWK) under the grant AuReSi (KK5335501JR1).

{\small
\bibliographystyle{ieee_fullname}
\bibliography{egpaper_final}
}

\end{document}